\documentclass[10pt,twocolumn,letterpaper]{article}

\usepackage[pagenumbers]{cvpr} % To force page numbers, e.g. for an arXiv version

\usepackage[dvipsnames]{xcolor}

\definecolor{cvprblue}{rgb}{0.21,0.49,0.74}
\usepackage{microtype}
\usepackage{graphicx}
\usepackage[pagebackref,breaklinks,colorlinks,citecolor=cvprblue]{hyperref}
\usepackage{multirow} 
\usepackage{amsmath}
\usepackage{amssymb}
\usepackage{mathtools}
\usepackage{amsthm}
\usepackage[capitalize,noabbrev]{cleveref}
\usepackage{subcaption}
\usepackage{graphicx}

\def\paperID{2} % *** Enter the Paper ID here
\def\confName{CVPR}
\def\confYear{2024}

\title{Exploring the Impact of Dataset Bias on Dataset Distillation}

\author{
Yao Lu\textsuperscript{1}\thanks{yaolu.zjut@gmail.com} 
\quad Jianyang Gu\textsuperscript{2}
\quad Xuguang Chen\textsuperscript{1}
\quad Saeed Vahidian\textsuperscript{3}
\quad Qi Xuan\textsuperscript{1}\thanks{Corresponding author, xuanqi@zjut.edu.cn.}
\\
\textsuperscript{1}{Zhejiang University of Technology}
\quad \textsuperscript{2}{Zhejiang University} 
\quad \textsuperscript{3}{Duke University}
}

\begin{document}
\maketitle

\begin{abstract}
Dataset Distillation (DD) is a promising technique to synthesize a smaller dataset that preserves essential information from the original dataset. This synthetic dataset can serve as a substitute for the original large-scale one, and help alleviate the training workload. However, current DD methods typically operate under the assumption that the dataset is unbiased, overlooking potential bias issues within the dataset itself. To fill in this blank, we systematically investigate the influence of dataset bias on DD. To the best of our knowledge, this is the first exploration in the DD domain. Given that there are no suitable biased datasets for DD, we first construct two biased datasets, CMNIST-DD and CCIFAR10-DD, to establish a foundation for subsequent analysis. Then we utilize existing DD methods to generate synthetic datasets on CMNIST-DD and CCIFAR10-DD, and evaluate their performance following the standard process. Experiments demonstrate that biases present in the original dataset significantly impact the performance of the synthetic dataset in most cases, which highlights the necessity of identifying and mitigating biases in the original datasets during DD. Finally, we reformulate DD within the context of a biased dataset. Our code along with biased datasets are available at \url{https://github.com/yaolu-zjut/Biased-DD}.
\end{abstract} 
\section{Introduction}
\label{sec:intro}

Recently, Dataset Distillation (DD) has attracted widespread attention within the deep learning community due to its potential to alleviate data burden and enhance training efficiency. It was first introduced by Wang et al.~\cite{wang2018dataset}, with the objective of condensing a large dataset into a small, synthetic one such that models trained on the latter yield comparable performance. After that, lots of subsequent study~\cite{zhao2020dataset,cazenavette2022dataset,zhou2022dataset,liu2023dream,nguyen2021dataset,lu2023can,cui2023scaling,sajedi2023datadam,zhang2023accelerating,du2023minimizing} has proposed a series of methods to improve the performance of synthetic datasets, including gradient matching~\cite{zhao2020dataset,zhao2021dataset,kim2022dataset}, trajectory matching~\cite{cazenavette2022dataset,du2023minimizing,cui2023scaling,du2023sequential} and distribution matching approaches~\cite{zhao2023dataset,wang2022cafe,zhao2023improved,sajedi2023datadam}. Despite achieving significant improvements, existing DD methods usually operate under the presupposition that the dataset is unbiased, overlooking potential issues within the dataset itself. However, in reality, datasets can be fraught with various problems, such as bias~\cite{torralba2011unbiased,khosla2012undoing,fabbrizzi2022survey}, imbalance~\cite{hasib2020survey,devi2020review}, label noise~\cite{algan2021image,song2022learning}, and missing values~\cite{emmanuel2021survey,feng2024adapting}, which can significantly affect the reliability and effectiveness of machine learning models and algorithms trained on these datasets. So, what happens when DD encounters dataset issues? How do dataset issues affect DD? Yet, there hasn't been research (either empirical or theoretical) that can answer this question.

% \textcolor{blue}{So, what if the dataset itself has inherent biases~\cite{torralba2011unbiased,khosla2012undoing,fabbrizzi2022survey}? How does dataset bias affect DD?}
To fill in this blank, we aim to investigate the influence of dataset issues on DD. In this paper, we concentrate on dataset bias, which arises when unintended attributes (i.e., bias attributes) are highly correlated with the label attribute within the dataset. For example, many images labeled as “camel” may have a “desert” background, creating an unintentional correlation. In this way, models mistakenly associate “camel” with “desert” instead of learning the actual characteristics of a camel. In this case, the desert is a dataset bias.

First of all, we create two biased datasets for DD, named CMNIST-DD and CCFAR10-DD, following the instructions of Nam et al.~\cite{nam2020learning}. Each dataset consists of 6 training sets with varying biased ratios (0\%, 10\%, 50\%, 80\%, 95\% and 100\%) and 1 unbiased testing set. We hope that these datasets can facilitate subsequent analysis on biased DD. Then we use several representative DD methods~\cite{zhao2020dataset,zhao2021dataset,zhao2023dataset} to generate synthetic datasets on CMNIST-DD and CCIFAR10-DD and evaluate their performance with the default parameter setting in the original papers. Experimental results demonstrate that dataset bias does affect DD in most cases. Therefore, it is essential to consider potential biases in datasets during DD. In view of this, we further provide a mathematical definition of DD with biased datasets, which we termed “biased DD” below. Compared to vanilla DD~\cite{zhao2020dataset,cazenavette2022dataset,zhou2022dataset,liu2023dream,nguyen2021dataset}, which aims to generate a small synthetic dataset that preserves as much information as possible from the original dataset, biased DD emphasizes unbiased attributes instead of the whole samples while minimizing the impact of biased attributes. We leave the specific implementation of biased DD to future work.

% need \textcolor{red}{Despite the existence of biased datasets~\cite{nam2020learning,lee2021learning,kim2021biaswap}, their high rates of bias aren't suitable for a comprehensive study.}

% While our findings are preliminary, our approach opens up new possibilities for designing

% In this work, we assume that we do not have annotations on the bias attribute a in the training dataset since they are expensive and laborious to obtain.

In summary, we emphasize our contributions as follows: 
\begin{itemize}
\item[$\bullet$] We propose a novel distillation scenario: distill valid information of large biased training sets into small, synthetic ones. To the best of our knowledge, we are the first to consider dataset biases during DD.
\item[$\bullet$] We create two biased datasets, named CMNIST-DD and CCFAR10-DD, to establish a foundation for subsequent analysis and the design of future debiased DD methods.
\item[$\bullet$] Having obtained CMNIST-DD and CCFAR10-DD, we conduct comprehensive experiments on them and conclude that dataset biases can seriously affect the performance of DD in most cases, which urgently calls for bias mitigation strategies specifically tailored for DD. Besides, we redefine DD when distilling biased datasets and leave the specific implementation to future work.
% \item[$\bullet$] \textcolor{blue}{We draw a counterfactual conclusion that the matching loss of bias-conflicting samples is smaller than bias-aligned samples. Inspired by this finding, we develop a simple, yet effective method named XX to identify and mitigate biases during the process of DD. Datasets are publicly available at \url{http}.}
\end{itemize}

\section{Related Work}
In this section, we briefly overview various dataset issues and existing work on DD.

\textbf{Dataset Issues.} Despite deep learning has achieved remarkable success in various fields~\cite{radford2021learning,achiam2023gpt,ramesh2022hierarchical}, the datasets used to train these models contain many issues that cannot be ignored. One critical issue is dataset bias~\cite{torralba2011unbiased,khosla2012undoing,fabbrizzi2022survey,zhang2018examining}, which arises when unintended attributes are highly correlated with the label. Such bias can lead to models that trained on these datasets producing inaccurate or unfair predictions. Another critical issue is data imbalance~\cite{leevy2018survey,krawczyk2016learning,peng2019trainable}, where certain classes are overrepresented in the dataset, resulting in models skewed towards those majority classes and performing poorly on minority classes. Additionally, due to the expensiveness of the labeling process or difficulty of correctly classifying data (even for the experts), label noise~\cite{algan2021image,song2022learning} becomes another common problem, which severely degrades the generalization performance of models. Missing values is also a common issue often attributed to human error, machine error, etc., and can cause performance degradation and data analysis problems.

\textbf{Dataset Distillation}, a method of compressing large datasets into smaller ones to improve training efficiency, was initially introduced by Wang et al.~\cite{wang2018dataset}. After that, many subsequent studies have introduced various matching losses to improve the performance of synthetic datasets. For example, DC~\cite{zhao2020dataset}, DSA~\cite{zhao2021dataset} and IDC~\cite{kim2022dataset} are proposed to match gradients between synthetic and original samples. MTT~\cite{cazenavette2022dataset}, LCMat~\cite{shin2023loss}, FTD~\cite{du2023minimizing}, TESLA~\cite{cui2023scaling} and DATM~\cite{guo2023towards} introduce a trajectory matching paradigm to minimize the loss of training trajectories between synthetic and original datasets. Different from matching in parameter space, DM~\cite{zhao2023dataset}, CAFE~\cite{wang2022cafe}, IDM~\cite{zhao2023improved} and DataDAM~\cite{sajedi2023datadam} use the feature space as the match proxy, and CLoM~\cite{lu2023can} utilizes pre-trained models to enhance the performance and cross-architecture generalization of synthetic datasets. Besides, DD has found extensive applications across various domains, including continual learning~\cite{yang2023efficient,gu2023summarizing}, privacy protection~\cite{dong2022privacy,vinaroz2023differentially}, federated learning~\cite{xiong2023feddm,zhang2022dense} and recommender systems~\cite{sachdeva2022infinite,wang2023gradient}.
% To this end, we investigate how dataset bias affects DD. To the best of our knowledge, our study is the first to consider issues within datasets used by DD.

Despite existing studies have demonstrated the effectiveness of DD and its application across various fields, they all hinge on the assumption of an unbiased dataset. Our study is dedicated to exploring DD under dataset bias, a topic that stands orthogonal to, yet distinct from, existing study.

% \subsection{Bias Mitigation}
% \textbf{Debiasing approaches with prior knowledge.} Several approaches pre-define a certain bias type in advance and  for addressing it. For example, However, assuming certain bias types does not guarantee that these approaches can generalize to other types of bias. 

% \textbf{Debiasing approaches without prior knowledge.}
% \textcolor{blue}{To this end, recent approaches aim to identify bias-conflicting samples and re-weighting them without making prior assumptions about a certain bias type. Nam et al.~\cite{nam2020learning} }  
\section{Preliminaries}
In this section, we present the formulation of dataset bias (\cref{sec:Definition}) and vanilla DD (\cref{sec:Definition DD}).
% Then we elaborate on how biased datasets are constructed in \cref{sec:Dataset Preparation}.
\subsection{Definition of Dataset Bias}
\label{sec:Definition}
Dataset bias is a dataset problem that occurs when unintended bias attributes hold a substantial correlation with the target attribute within the training dataset. To be specific, suppose $x$ is a biased image sampled from a dataset with its corresponding label $y$, $z$ is the attributes extracted from $x$. Among these attributes, $z_g$ denotes the attribute that is essential for predicting a target label $y$, while $z_b$ denotes the attribute that is less essential, but has a strong correlation with $y$. Since $z_b$ is easier for the model to learn compared to $z_g$~\cite{nam2020learning}, the model becomes biased by overly exploiting $z_b$ instead of $z_g$ when trained on the biased dataset, failing to predict the samples which do not contain $z_b$.

For instance, many images labeled as “camel” may contain a “desert” background. This unintentional correlation will mislead models into associating “camel” with “desert” instead of learning the actual characteristics of a “camel”. Samples that have a strong correlation (like “camel in the desert”) are called \textbf{bias-aligned samples}, while samples that have a weak correlation (like “camel on the grass”) are termed \textbf{bias-conflicting samples}. Finally, the biased rate of the dataset can be calculated by \cref{eq1}, where $N_{ba}$ and $N_{bc}$ denote the number of bias-aligned samples and bias-conflicting samples respectively.
\begin{equation}
    \textit{Biased Rate}=\frac{N_{ba}}{N_{bc}+N_{ba}}
    \label{eq1}
\end{equation}

% \textcolor{green}{In order to ensure that the classifier has good generalization performance, it must include attributes that are strongly correlated with the target label. However, if the classifier were more likely to learn information related to unimportant attributes, so that it can be biased. For instance, we consider camel as the target which we need to classify. In our daily lives, camel always occurs on the desert, so we give the classifier a lots of images about camel on the desert. The model then becomes biased when we give a image that is the camel on the grass.} 

\begin{figure*}[t]
	\centering
	\begin{minipage}[t]{0.49\linewidth}
		\centering
        \subfloat[\label{fig:a}]{\includegraphics[width=0.99\textwidth]{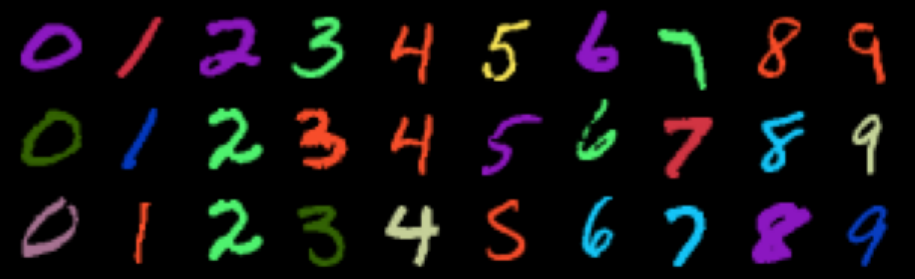}} \\
        \vspace{1mm}
		\subfloat[\label{fig:b}]{\includegraphics[width=0.99\textwidth]{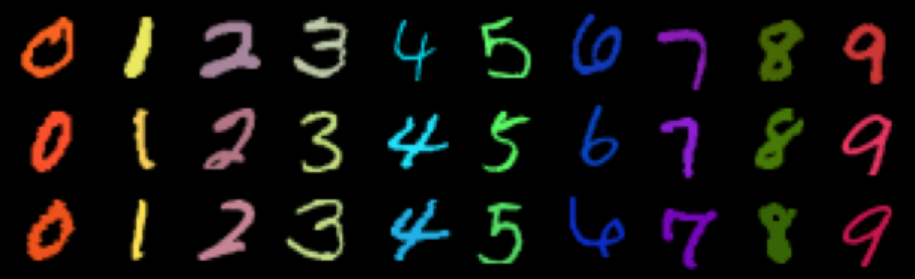}} \\
        \vspace{1mm}
        \subfloat[\label{fig:c}]{\includegraphics[width=0.99\textwidth]{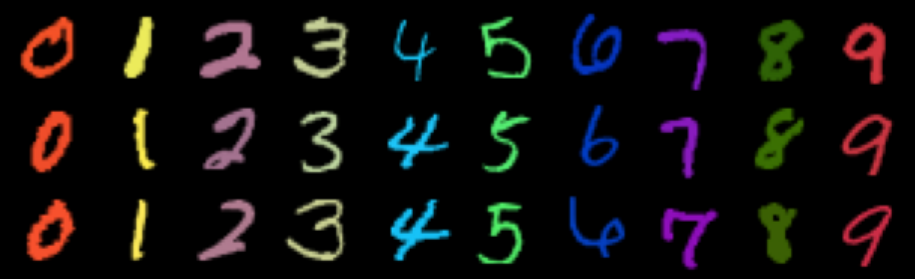}} \\
        \vspace{1mm}
        \subfloat[\label{fig:d}]{\includegraphics[width=0.99\textwidth]{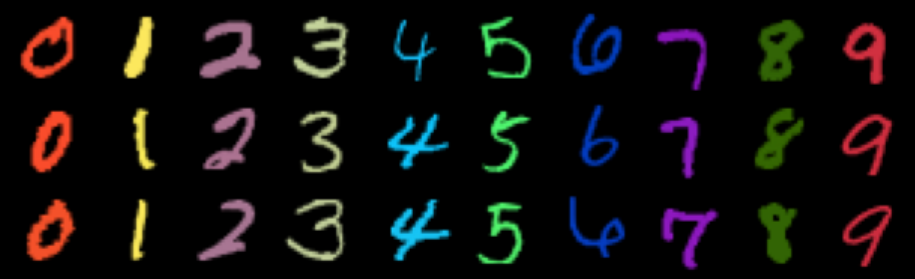}} \\
        \vspace{1mm}
        \subfloat[\label{fig:e}]{\includegraphics[width=0.99\textwidth]{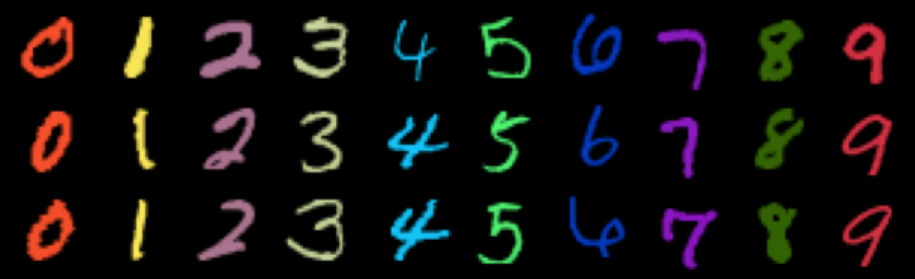}} \\
        % \caption{CMNIST-DD, severity=1,2,3,4}
	\end{minipage}
	\begin{minipage}[t]{0.49\linewidth}
		\centering
        \subfloat[\label{fig:f}]{\includegraphics[width=0.99\textwidth]{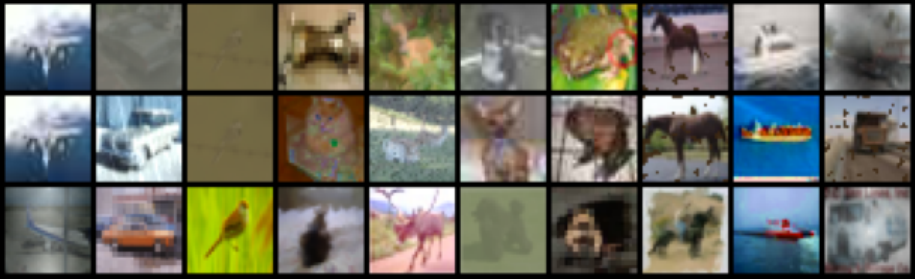}} \\
        \vspace{1mm}
		\subfloat[\label{fig:g}]{\includegraphics[width=0.99\textwidth]{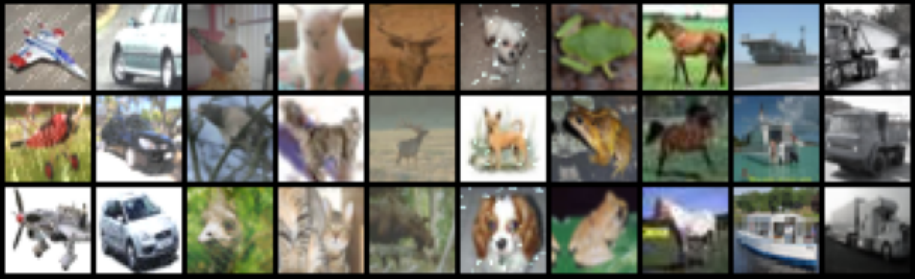}} \\
        \vspace{1mm}
        \subfloat[\label{fig:h}]{\includegraphics[width=0.99\textwidth]{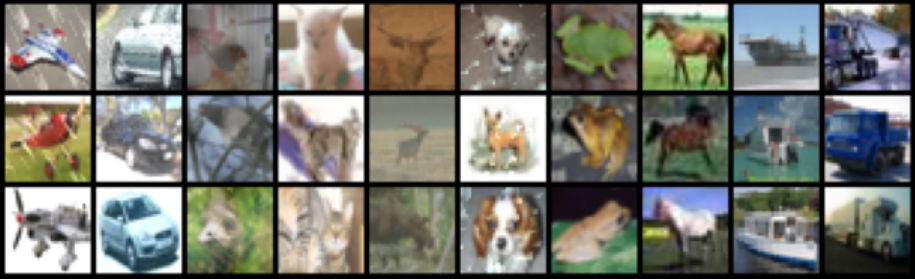}} \\
        \vspace{1mm}
        \subfloat[\label{fig:i}]{\includegraphics[width=0.99\textwidth]{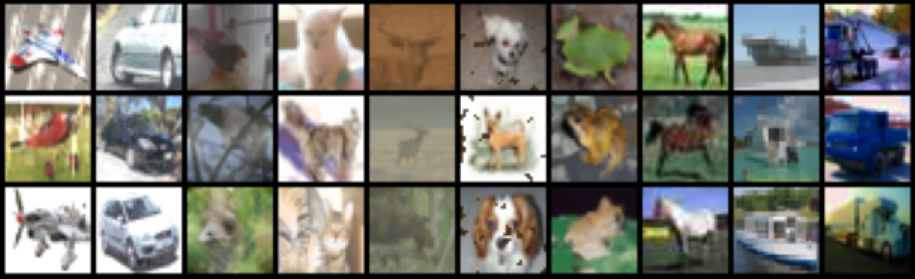}} \\
        \vspace{1mm}
        \subfloat[\label{fig:j}]{\includegraphics[width=0.99\textwidth]{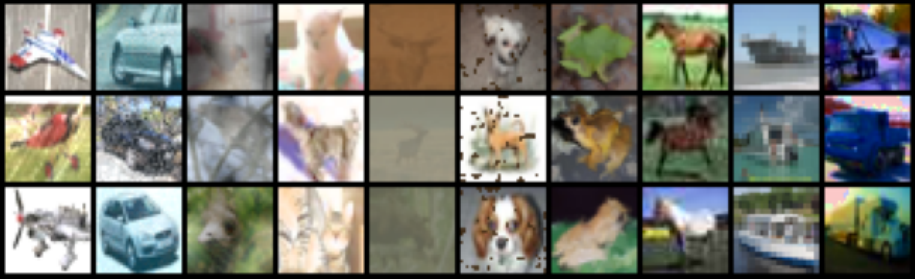}} \\
        % \caption{CCIFAR10-DD, severity=1,2,3,4}
	\end{minipage}
 \caption{Visualizations of bias-conflicting samples and bias-aligned samples. Figure (a) and (f) visualize the bias-conflicting samples in CMNIST-DD and CCIFAR10-DD, respectively. Figure (b)-(e) and (g)-(j) visualize the bias-aligned samples with various severities in CMNIST-DD and CCIFAR10-DD, respectively. Severity increases from top to bottom. As for CCIFAR10-DD, we add 10 types of corruptions to 10 categories of CIFAR10. Specifically, “snow” for “airplane”, “frost” for “automobile”, “fog” for “bird”, “brightness” for “cat”, “contrast” for “deer”, “spatter” for “dog”, “elastic” for “frog”, “JPEG” for “horse”, “pixelate” for “ship” and “saturate” for “truck”. Best viewed in color.}
 \label{fig:biased samples}
 \vspace{-3mm}
\end{figure*}

\subsection{Definition of Vanilla DD}
\label{sec:Definition DD}
Vanilla DD is built on the assumption of an unbiased dataset. Its goal is to generate a small synthetic dataset that retains as much information as possible from the original dataset.

Assume that we are given a large training set $\mathcal{T} = \{(x_i, y_i)\}|^{|\mathcal{T}|}_{i=1}$, the synthetic dataset $\mathcal{S} = \{(\hat{x_i}, y_i)\}|^{|\mathcal{S}|}_{i=1}$ ($|\mathcal{S}| \ll |\mathcal{T}|$), generated by vanilla DD, can be obtained by solving the following minimization problem:
\begin{equation}
\min _{\mathcal{S}} \mathcal{D}(\mathcal{S}, \mathcal{T}),
\label{eq2}
\end{equation}
where $\mathcal{D}$ is a task-specific matching loss. 

% However, in real-world scenarios, datasets may be fraught with dataset bias issues~\cite{torralba2011unbiased,khosla2012undoing,fabbrizzi2022survey}. Considering the existence of bias in the dataset, we reformulate the problem as follows:

% Let $\mathcal{T}_b = \{(x_{j}, z_{g,j}, z_{b,j}, y_j)\}|_{j=1}^{|\mathcal{T}_b|}$ be the set of bias-aligned samples, where $x_{j}$ and $y_j$ are the $j$-th bias-aligned sample and its label, $z_{g,j}$ and $z_{b,j}$ denote its unbiased attribute and biased attribute. Similarly, the set of bias-conflicting samples can be formulated as $\mathcal{T}_g = \{(x_{k}, z_{g,k}, y_k)\}|_{k=1}^{|\mathcal{T}_g|}$, and $|\mathcal{T}_b| + |\mathcal{T}_g| = |\mathcal{T}|$. The objective of Biased DD is to extract and retain the unbiased attributes $z_g$ from both biased set $\mathcal{T}_b$ and unbiased set $\mathcal{T}_g$, while minimizing the impact of biased attributes $z_b$. Specifically, it can be formalized as the following optimization problem:
% \begin{equation}
% \min _{\mathcal{S}} \mathcal{D}(\mathcal{S}, \mathcal{A}_g) - \lambda \mathcal{D}(\mathcal{S}, \mathcal{A}_b),
% \end{equation}
% where $\mathcal{A}_g$ is the composite of unbiased attributes present in both $\mathcal{T}_b$ and $\mathcal{T}_g$, $\mathcal{A}_b$ is the collection of biased attributes within $\mathcal{T}_b$. $\lambda$ is a regularization balancing the contribution of unbiased and biased attributes in the optimization process.

% $\mathcal{T}_b = \{(x_{b1}, y_1), (x_{b2}, y_2), \cdots, (x_{|\mathcal{T}_b|}, y_{|\mathcal{T}_b|})\}$}, 

\section{Dataset Bias in Synthetic Datasets}
\label{sec:Dataset Bias}

\begin{figure*}[t]
	\centering
    \centering
    \includegraphics[width=0.97\textwidth]{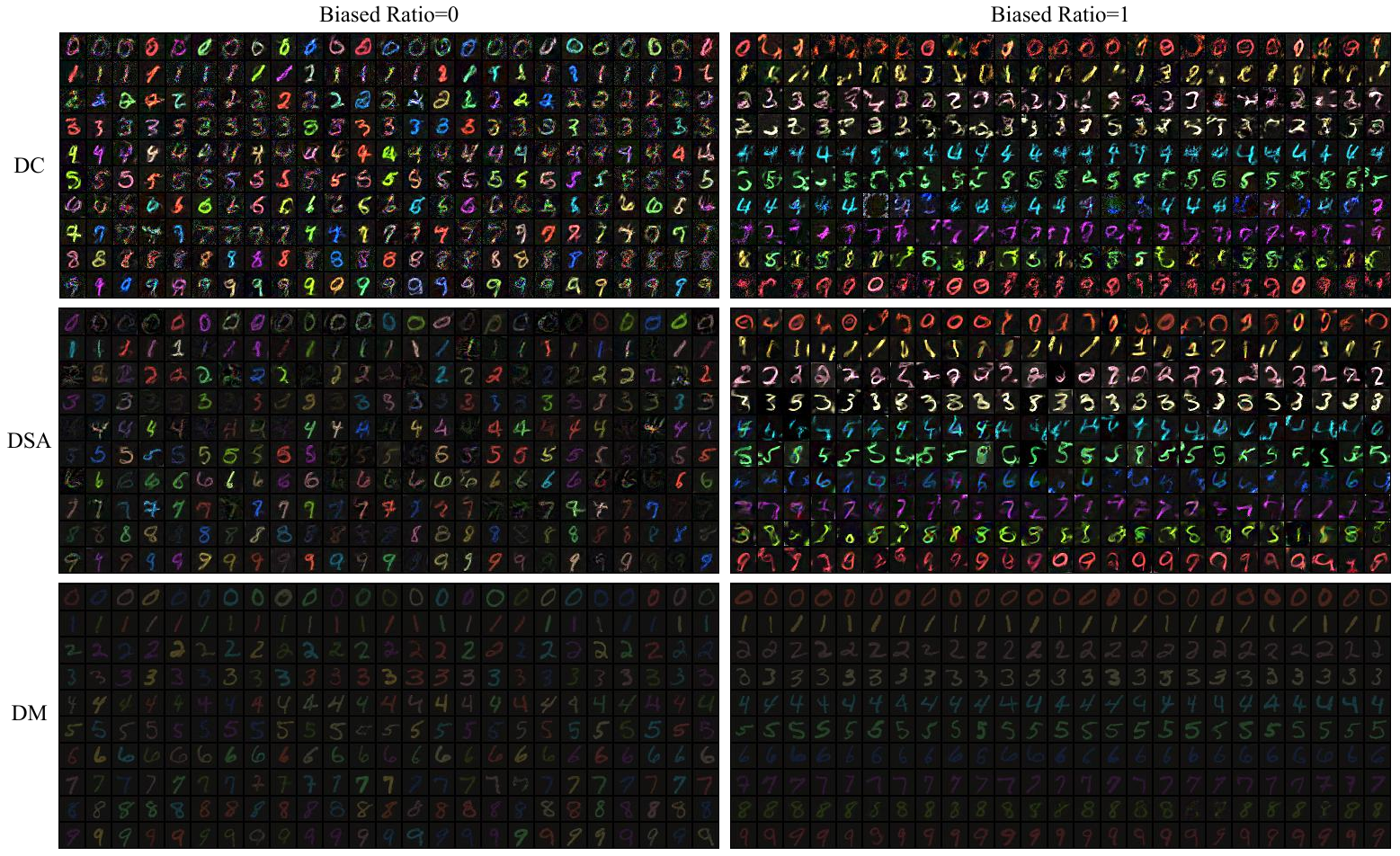} 
 \caption{Visualizations of synthetic datasets generated by various DD methods on CMNIST-DD. All experiments are conducted at a severity level of 4.}
 \label{fig:vis dd severity=4 CMNIST-DD}
\end{figure*}

\begin{figure*}[t]
	\centering
    \centering
    \includegraphics[width=0.97\textwidth]{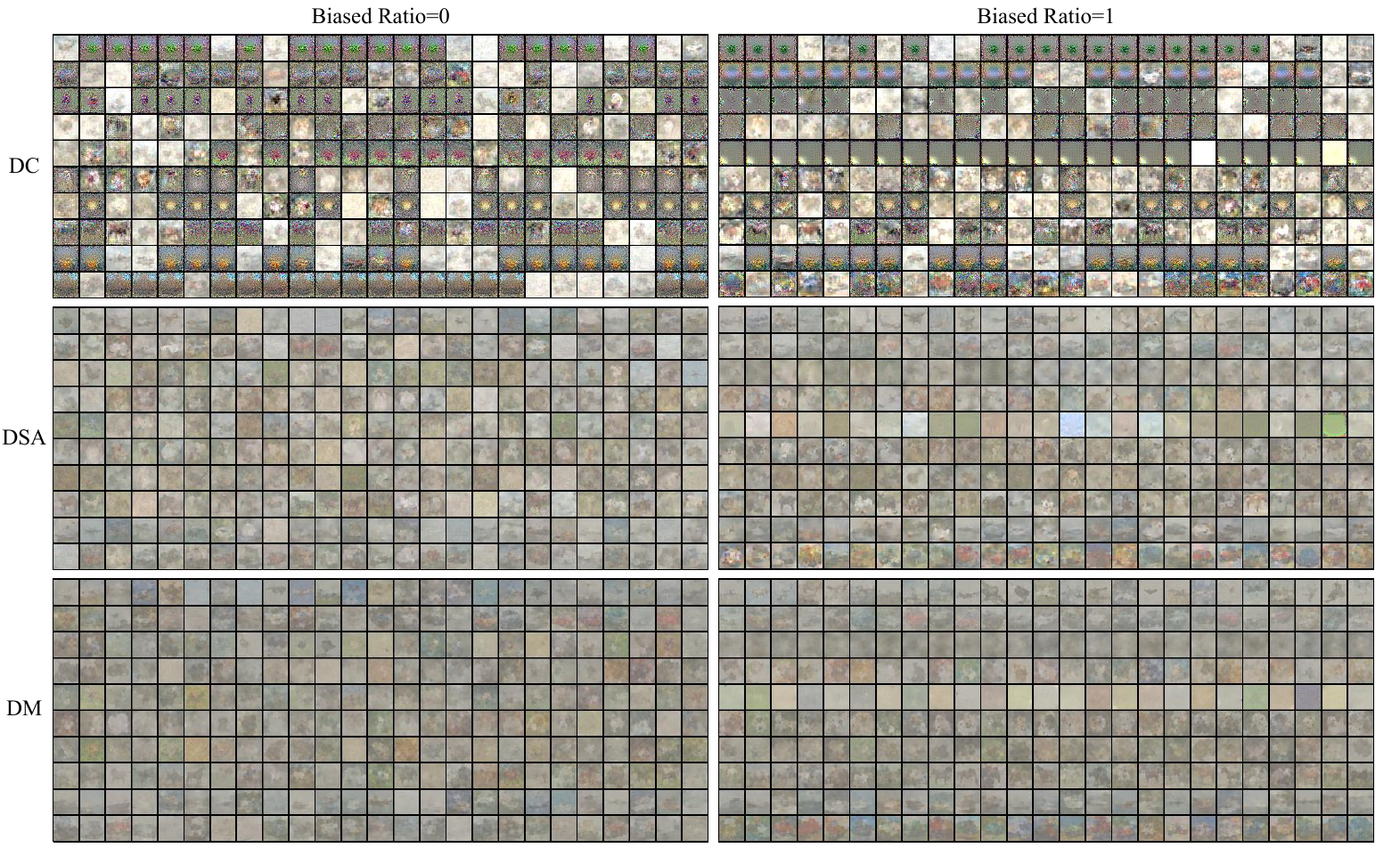} 
 \caption{Visualizations of synthetic datasets generated by various DD methods on CCIFAR10-DD. All experiments are conducted at a severity level of 4.}
 \label{fig:vis dd severity=4 CCIFAR10-DD}
 \vspace{-2mm}
\end{figure*}

\subsection{Biased Dataset Preparation}
\label{sec:Dataset Preparation}
Although biased datasets such as Colored MNIST and Corrupted CIFAR10~\cite{nam2020learning} already exist, these datasets exhibit high levels of bias (biased ratio: $95.0\%$, $98.0\%$, $99.0\%$ and $99.5\%$), which is not conducive to comprehensively analyzing the impact of dataset bias on DD. To this end, we construct two biased datasets, CMNIST-DD and CCFAR10-DD, following the instructions of Nam et al.~\cite{nam2020learning}. Specifically, as for CMNIST-DD, we select ten distinct colors and inject each color with random perturbation into the foreground of each digit of MNIST~\cite{lecun1998gradient}. By adjusting the number of bias-aligned samples in the training set, we obtain six different datasets with the ratio of bias-aligned samples of 0\%, 10\%, 50\%, 80\%, 95\% and 100\%. As for CCFAR10-DD, we utilize a set of protocols~\cite{hendrycks2019benchmarking} for corruption and inject them into CIFAR10~\cite{krizhevsky2009learning}. Specifically, “snow” for “airplane”, “frost” for “automobile”, “fog” for “bird”, “brightness” for “cat”, “contrast” for “deer”, “spatter” for “dog”, “elastic” for “frog”, “JPEG” for “horse”, “pixelate” for “ship” and “saturate” for “truck”. CCFAR10-DD also has six different datasets with their correlation ratios as in CMNIST-DD. Finally, a parameter, \textit{severity}, is introduced to regulate the intensity of disturbance on CMNIST-DD and CCFAR10-DD datasets. \cref{fig:biased samples} exhibits the bias-aligned samples under severity=1-4 for CMNIST-DD and CCFAR10-DD.

\subsection{Experimental Setups}
Having obtained CMNIST-DD and CCFAR10-DD, we next utilize them as the original dataset and perform DD on them. In this paper, we select three representative DD methods for experiments: gradient-matching based DC~\cite{zhao2020dataset}, DSA~\cite{zhao2021dataset} and distribution-matching based DM~\cite{zhao2023dataset}.

Specifically, DC aligns the training gradients derived from synthetic samples with those obtained from original samples. Given a model with parameters $\theta$, the optimization process can be expressed as:
\begin{equation}
    \min_{\mathcal{S}} \sigma\left(\nabla_\theta\mathcal{L}(\theta;\mathcal{S}),\nabla_\theta\mathcal{L}(\theta;\mathcal{T})\right),
\end{equation}
where $\mathcal{L}(\cdot;\cdot)$ denotes the training loss and $\sigma(\cdot;\cdot)$ represents the distance measure. On the basis of DC, DSA further applies data augmentation techniques to improve the performance of synthetic datasets:
\begin{equation}
\min_{\mathcal{S}} \sigma\left(\nabla_{\theta}\mathcal{L}(\mathcal{A}(\mathcal{S},\omega^{\mathcal{S}}), \theta),\nabla_{\theta}\mathcal{L}(\mathcal{A}(\mathcal{T},\omega^{\mathcal{T}}), \theta)\right),
\end{equation}
where ${\mathcal{A}}$ is a family of image transformations such as cropping, color jittering and flipping that are parameterized with $\omega^{\mathcal{S}}$ and $\omega^{\mathcal{T}}$ for synthetic and real training sets respectively. DM aligns the feature distributions of synthetic and real training sets using maximum mean discrepancy~\cite{gretton2012kernel} in sampled embedding spaces:
\begin{equation}
    \min_{\mathcal{S}} \Vert\frac{1}{|\mathcal{T}|}\sum_{i=1}^{|\mathcal{T}|}f(\theta;\mathcal{A}(x_i,\omega^{\mathcal{T}}))-\frac{1}{|\mathcal{S}|}\sum_{i=1}^{|\mathcal{S}|}f(\theta;\mathcal{A}(s_i,\omega^{\mathcal{S}}))\Vert^2,
\end{equation}
where $f(\cdot;\cdot)$ is the feature extraction function.

\textbf{Implementation Details.} In this paper, we use the default hyperparameter settings of DC, DSA and DM\footnote{https://github.com/VICO-UoE/DatasetCondensation} to synthesize datasets and evaluate their performance. As for evaluating the original CMNIST-DD and CCIFAR10-DD, we set the batch size, weight decay, epoch and momentum to 256, 0.0005, 150 and 0.9, respectively. The optimizer is set as SGD, with an initial learning rate of 0.01. The learning rate is decayed by a factor of 0.1 at epochs $50$ and $100$. Besides, we repeat each experiment $3$ times and report the mean and standard deviation.

\begin{table*}[t]
  \centering
    \begin{tabular}{cp{7.5em}cccccc}
    \toprule
    \multirow{2}[2]{*}{Dataset} & \multicolumn{1}{c}{\multirow{2}[2]{*}{Method}} & \multicolumn{6}{c}{Ratio of biaseds-aligned samples} \\
          & \multicolumn{1}{c}{} & 0\%   & 10\%  & 50\%  & 80\%  & 95\% & 100\%\\
    \midrule
    \multirow{3}[2]{*}{CMNIST-DD} & \multicolumn{1}{c}{Full set} &  \textbf{99.49$\pm$0.02}     &  \textbf{99.49$\pm$0.02}     &  \textbf{99.29$\pm$0.01 }    &  \textbf{98.50$\pm$0.05}    & \textbf{95.22$\pm$0.11}  & \textbf{8.89$\pm$0.50} \\
          & \multicolumn{1}{c}{DC} & 97.33$\pm$0.11      & 97.72$\pm$0.12& 93.77$\pm$0.30 &  86.87$\pm$1.20 &65.10$\pm$2.26 & 0.23$\pm$0.18\\
          & \multicolumn{1}{c}{DSA} & 98.08$\pm$0.10   &98.18$\pm$0.06 & 97.26$\pm$0.04 & 95.34$\pm$0.10  & 84.29$\pm$0.90 &1.44$\pm$0.14 \\
          & \multicolumn{1}{c}{DM} &  97.42$\pm$0.02     &97.32$\pm$0.04 & 94.03$\pm$0.32 & 74.15$\pm$0.09 & 12.92$\pm$0.53 & 6.45$\pm$0.91\\
    \midrule
    \multirow{3}[2]{*}{CCIFAR10-DD}  & \multicolumn{1}{c}{Full set} &   \textbf{73.77$\pm$0.35}    &   \textbf{72.86$\pm$0.12}    &  \textbf{67.13$\pm$0.36}     &   \textbf{55.02$\pm$0.18}    & 37.34$\pm$0.36  &24.55$\pm$0.72\\
          & \multicolumn{1}{c}{DC} &  41.68$\pm$0.17     &41.57$\pm$0.47 & 36.76$\pm$0.43 & 29.70$\pm$0.11 & 27.61$\pm$0.36& 25.99$\pm$0.48 \\
          & \multicolumn{1}{c}{DSA} & 49.52$\pm$0.29      &   48.72$\pm$0.50    &   42.86$\pm$0.35    &   37.45$\pm$0.44    & 34.24$\pm$0.61  & 33.31$\pm$0.46\\
          & \multicolumn{1}{c}{DM} &  52.29$\pm$0.51     &  52.54$\pm$0.25     &   47.47$\pm$0.73    &  42.13$\pm$0.21    &  \textbf{38.23$\pm$0.47} & \textbf{36.68$\pm$0.15} \\
    \bottomrule
    \end{tabular}%
     \caption{Performance of synthetic datasets (IPC=50) generated by different DD methods on CMNIST-DD and CCIFAR10-DD with varying ratios of bias-aligned samples (severity=4). Performance is evaluated on unbiased samples. “Full sets” means the model is trained on the original full dataset without distillation. \textbf{Bold entries} are best results.}
  \label{tab:1}%
\end{table*}%

\begin{table}[t]
  \centering
    \scalebox{0.7}{
    \begin{tabular}{ccccc}
    \toprule
    Severity & 1     & 2     & \multicolumn{1}{c}{3} & 4 \\
    \midrule
    CMNIST-DD BR=0  & 97.28$\pm$0.09 &  97.46$\pm$0.07     &  97.41$\pm$0.01     & 97.42$\pm$0.02 \\
    \midrule
    CMNIST-DD BR=100 & 8.65$\pm$0.24 &  7.78$\pm$1.02     & 6.60$\pm$0.36 & 6.45$\pm$0.91 \\
    \midrule
    CCIFAR10-DD BR=0  & 57.40$\pm$0.45 &  55.75$\pm$0.37     &   56.04$\pm$0.15    & 52.29$\pm$0.51 \\
    \midrule
    CCIFAR10-DD BR=100 & 49.92$\pm$0.10 &  50.06$\pm$0.44    & 45.10$\pm$0.13 & 36.68$\pm$0.15 \\
    \bottomrule
    \end{tabular}}
\caption{The effect of perturbation severity on the performance of synthetic datasets. BR denotes biased ratio. All experiments are conducted using DM, with 50 images per class.}\
  \label{tab:Influence of severity}%
  \vspace{-6mm}
\end{table}%

\subsection{Experiments}
We use the default hyperparameter settings of DC, DSA and DM to synthesize datasets (50 images per class) on CMNIST-DD and CCIFAR10-DD. Specifically, we use datasets with biased ratios of 0 and 1, at a severity level of 4 to conduct experiments. For clearer visualization, we select a subset of samples from the synthetic dataset and visualize them. \cref{fig:vis dd severity=4 CMNIST-DD} and \cref{fig:vis dd severity=4 CCIFAR10-DD} exhibit the visualizations of synthetic datasets generated by various DD methods on CMNIST-DD and CCIFAR10-DD, respectively. We find that when the biased ratio is 0, digits in the same class have completely different colors. However, when the biased ratio is 1, digits in the same class share the same color, which reveals that the color attribute has indeed been encoded into the synthetic datasets as a significant feature. As for CCIFAR10-DD, synthetic datasets generated from the biased dataset exhibit less diversity and richness compared to those derived from the unbiased dataset. We believe this phenomenon can be attributed to the biases in the original datasets that skew the distribution of features.

After that, we generate corresponding synthetic datasets using datasets with different biased ratios ($0\% $, $10\%$, $50\%$, $80\%$, $95\%$, $100\%$) and evaluate their performance on an unbiased test set. Furthermore, we evaluate the performance of the model trained on the original dataset with various biased ratios as a control. As shown in \cref{tab:1}, we observe that when the biased ratio is relatively low, the performance impact on synthetic CMNIST-DD is minimal. In other words, the performance of the synthetic dataset is relatively similar to that of the original CMNIST-DD. However, as the biased ratio increases ($\geq 50\%$), the performance disparity between the synthetic dataset and original CMNIST-DD is gradually increasing, which means DD is affected by dataset biases. When the biased ratio reaches $100\%$, the performance on the synthetic dataset experiences a dramatic decline, but the performance gap between the synthetic dataset and the original CMNIST-DD narrows significantly. As for CCIFAR10-DD, when the biased ratio is below $80\%$, the performance gap between the synthetic and original datasets is quite large, which means dataset biases do affect DD. However, when the biased ratio exceeds $95\%$, the performance of the synthetic dataset even higher than that of the original dataset. This phenomenon indicates that DD can retain more useful information of the original dataset under extreme bias rates (nearly $100\%$), offering a brand new perspective into the design of dataset debiasing. 
% In contrast, dataset biases have a significant impact on DD, which highlights the necessity of considering bias during DD and the urgency of designing debiased DD methods.

To delve into the impact of severity on DD, we conduct a series of experiments utilizing DM. By adjusting the severity ($1$-$4$) of the biased training set, we generate corresponding synthetic datasets and subsequently evaluate their performance on an unbiased test set. As illustrated in \cref{tab:Influence of severity}, the severity of disturbance also has a notable impact on the performance of synthetic datasets. Specifically, biased datasets are more susceptible to the increased disturbance severity, resulting in a more significant performance degradation than their unbiased counterparts.

In summary, although DD is less affected by dataset bias or even benefits from it at low and very high bias rates, but in most cases, dataset bias considerably impacts DD, which highlights the necessity of identifying and mitigating biases in the original datasets during DD.

% \subsection{Loss}
% \textcolor{red}{Next, we observe that the loss dynamics of bias-aligned samples and bias-conflicting samples during the training phase are in stark contrast. This observation serves as a key intuition for designing our debiasing algorithm.}

\section{Biased DD}
In the previous section, we have demonstrated that dataset bias does affect DD and vanilla DD methods fail when faced with biased datasets in most cases, which indicates that the vanilla definition of DD is no longer suitable for biased datasets. To this end, we reformulate DD within the context of biased datasets, which we call biased DD, as follows:
% However, in real-world scenarios, datasets may be fraught with dataset bias issues~\cite{torralba2011unbiased,khosla2012undoing,fabbrizzi2022survey}. Considering the existence of bias in the dataset, we reformulate the problem as follows:

Let $\mathcal{T}_b = \{(x_{j}, z_{g,j}, z_{b,j}, y_j)\}|_{j=1}^{|\mathcal{T}_b|}$ be the set of bias-aligned samples, where $x_{j}$ and $y_j$ are the $j$-th bias-aligned sample and its label, $z_{g,j}$ and $z_{b,j}$ denote its unbiased attribute and biased attribute. Similarly, the set of bias-conflicting samples can be formulated as $\mathcal{T}_g = \{(x_{k}, z_{g,k}, y_k)\}|_{k=1}^{|\mathcal{T}_g|}$, and $|\mathcal{T}_b| + |\mathcal{T}_g| = |\mathcal{T}|$. The objective of biased DD is to extract and retain the unbiased attributes $z_g$ from both biased set $\mathcal{T}_b$ and unbiased set $\mathcal{T}_g$, while minimizing the impact of biased attributes $z_b$. Specifically, it can be formalized as the following optimization problem:
\begin{equation}
\min _{\mathcal{S}} \mathcal{D}(\mathcal{S}, \mathcal{A}_g) - \lambda \mathcal{D}(\mathcal{S}, \mathcal{A}_b),
\end{equation}
where $\mathcal{A}_g$ is the composite of unbiased attributes present in both $\mathcal{T}_b$ and $\mathcal{T}_g$, $\mathcal{A}_b$ is the collection of biased attributes within $\mathcal{T}_b$. $\lambda$ is a regularization balancing the contribution of unbiased and biased attributes in the optimization process. We will leave the specific implementation of biased DD to future work.

% \vspace{-2mm}
\section{Conclusion}
% \vspace{-1mm}
In this paper, we delve into DD when the dataset is endogenously biased. To the best of our knowledge, this is the first exploration in the DD field. Specifically, we construct two biased datasets, namely CMNIST-DD and CCIFAR10-DD and conduct a series of experiments on them. Experimental results show that dataset biases indeed influence DD in most cases, highlighting the necessity of designing bias mitigation strategies specifically tailored for DD. Therefore, we propose a mathematical definition of biased DD and leave the specific implementation to future research.

\textbf{Future Work.} In this paper, we have demonstrated that dataset bias does affect DD. In the future, we will extend our experiments to larger datasets, more complex models and more advanced DD methods. Besides, it would be interesting to investigate why synthetic datasets outperform original datasets under extreme bias rates and what can be done with this phenomenon. Furthermore, how to eliminate or mitigate the impact of biased samples on synthetic datasets during the DD process is also a promising direction.
% we emphasize the importance of acknowledging biases within datasets during DD and stress the necessity of creating bias mitigation strategies specifically tailored for DD

% We advocate for recognizing the biases present within datasets during distillation and highlight the pressing need for developing bias mitigation strategies specifically designed for DD.

% \textbf{Future Work.}  A promising direction is debias .

\clearpage
{
    \small
    \bibliographystyle{ieeenat_fullname}
    \bibliography{main}
}


\end{document}